\documentclass{article}
\usepackage{stywhispers,amsmath,epsfig}
\usepackage{graphicx}
\usepackage[table]{xcolor}
\usepackage{stfloats}
\usepackage{afterpage}
\usepackage{url}
\usepackage{hyperref}
\hypersetup{
    colorlinks=true,
    linkcolor=black,
    filecolor=black, 
    citecolor=black,
    urlcolor=blue,
    }
\usepackage{varwidth}

\title{Masking Hyperspectral Imaging Data With Pretrained Models}
\name{\parbox{\linewidth}{\centering \normalfont Elias Arbash, Andréa de Lima Ribeiro, Sam Thiele, Nina Gnann, \\ Behnood Rasti, Margret Fuchs, Pedram Ghamisi, Richard Gloaguen}}

\address{Helmholtz-Zentrum Dresden-Rossendorf, \\ Helmholtz Institute Freiberg for Resource Technology, Freiberg, Germany}
\begin{document}
%\ninept
%
\maketitle
\begin{abstract}
The presence of undesired background areas associated with potential noise and unknown spectral characteristics degrades the performance of hyperspectral data processing. Masking out unwanted regions is key to addressing this issue. Processing only regions of interest yields notable improvements in terms of computational costs, required memory, and overall performance.
The proposed processing pipeline encompasses two fundamental parts: regions of interest mask generation, followed by the application of hyperspectral data processing techniques solely on the newly masked hyperspectral cube. The novelty of our work lies in the methodology adopted for the preliminary image segmentation. We employ the Segment Anything Model (SAM) to extract all objects within the dataset, and subsequently refine the segments with a zero-shot Grounding Dino object detector, followed by intersection and exclusion filtering steps, without the need for fine-tuning or retraining. To illustrate the efficacy of the masking procedure, the proposed method is deployed on three challenging applications scenarios that demand accurate masking i.e. shredded plastics characterization, drill core scanning, and litter monitoring. 
The numerical evaluation of the proposed masking method on the three applications is provided along with the used hyperparameters. The scripts for the method will be available at \url{https://github.com/hifexplo/Masking}.
\end{abstract}
\section{Introduction}
\label{sec:intro}

The deployment of Machine Learning (ML) has significantly improved task efficiency and performance across various domains, e.g.,  medical field, agriculture, industry, and remote sensing \cite{intro}. Deep learning, a subfield of ML, focuses on utilizing neural networks to approximate complex functions present in the data. However, supervised learning scenarios often necessitate significant amounts of labeled data for training deep learning models. The process of data labeling can be labor-intensive and time-consuming, particularly for computer vision segmentation tasks that require precise labeling. Moreover, numerous real-world applications demand accurate segmentation even in the absence of large-scale training datasets.
The number of publicly available hyperspectral datasets together with their detailed labeled samples are limited \cite{9884725}. 
This pressing shortcoming in the hyperspectral community usually hinders the optimization of deep learning models for downstream tasks such as classification and segmentation.
The existence of undesirable backgrounds can obstruct the segmentation performance. In tasks where the scene contains a large area of background along the objects of interest, the performance of classification techniques is compromised at multiple stages of the processing pipeline, e.g:
\begin{itemize}
    \item Dimensionality reduction: methods like Principal Component Analysis (PCA) (band-wise) build upon calculating the mean of the values in the hyperspectral scene. Having a large area of irrelevant background means having many pixels skewing the calculation, yielding impaired desired-objects representative principal components, i.e., exerting a negative impact on the new representation of the object of interest in the hyperspectral scene. Excluding the background from the calculations enhances the indicative principal components.
    \item  Machine learning models: during an RGB segmentation task, the model can focus on learning the visual characteristics of the objects and differentiate them from the surrounding regions implicitly since objects are distinct in terms of color, texture, or shape. However, having an irrelative background in the hyperspectral scene can exhibit significant spectral variations due to illumination changes, shadows, sensor deficiencies, or different materials present in the scene. This leads to noisy predictions in the model's output. This can be problematic in tasks where the spatial coordination of the segmented pixels is needed. Therefore, masking out unwanted backgrounds can help in this manner.
\end{itemize}

To address these challenges and improve model performance in scenarios with limited training data, we propose a segmentation method that preserves the objects of interest and allows the exclusion of the background without the need for any retraining or fine-tuning. The method's pipeline leverages the Segment Anything Model (SAM); developed by Meta \cite{sam} and Grounding Dino zero-shot object detector \cite{dino}.

The remainder of this paper is structured as follows: Section two provides a detailed explanation of the methodology and its components. Section three is a discussion of the three applications on which the method was tested. We present the numerical evaluation of its segmentation performance, the results obtained, the hyperparameters used, and suggestions for subsequent hyperspectral processing techniques. Section four focuses on the discussion of masking methods, while section five concludes the work.

\section{Methodology}
\label{sec: methodology}

A key factor in building models with good segmentation performance is the availability of large training datasets. However, in the hyperspectral community, this is not feasible due to the limitation of publicly available hyperspectral datasets. Therefore, our work is motivated by the use of pre-trained models, specifically the utilization of SAM and the zero-shot grounding dino object detector.

\subsection{Segment Anything Model (SAM)}

SAM is an advanced segmentation model that builds upon incorporating several key modifications to enhance efficiency and scalability. It uses a masked auto-encoder pre-trained vision transformer with minimal adaptations to handle high-resolution inputs, a prompt encoder to encode sparse and dense prompts into an embedding vector in real time, and a lightweight mask decoder that predicts the segmentation masks based on embeddings from both the image and prompt encoder\cite{sam}. The model has been trained on an extensive dataset comprising 1.1 billion masks and 1 million images from diverse sources \cite{sam}. 
Experimental results have demonstrated the effectiveness of SAM, showcasing superior performance compared to existing state-of-the-art methods on benchmark datasets. It serves as a comprehensive segmentation model capable of segmenting all objects within an input image \cite{zhang2023comprehensive}. However, it should be noted that SAM's segmentation does not include object identification or labeling, which may pose limitations for direct deployment in specific tasks.

To address this limitation and refine the segmentations predicted by SAM, the zero-shot grounding dino object detection model is employed. This additional model focuses on desired targets or objects of interest, allowing for more precise and tailored segmentation \cite{dino}.

\subsection{Grounding Dino}
Grounding dino is a novel zero-shot object detector that leverages user-provided descriptive textual prompts to accurately detect the specified objects of interest with a certain confidence threshold \cite{dino}, which can be used to emphasize the desired masks predicted by SAM. Grounding dino leverages the transformer architecture, to facilitate the comprehension of visual input through the association of textual and visual information\cite{dino}. 

Incorporating SAM and grounding dino into one processing pipeline, along an intersection, and excluding filtering steps, non-desired abundant segmentation masks generated by SAM can be filtered out effectively.  It allows accurate segmentation masking for objects of interest without fine-tuning or retraining needed.
Through the combined utilization of SAM and grounding dino, we achieve the effective removal of undesired segmentation masks generated by SAM. The method pipeline can be stated with the following steps:
\begin{enumerate}
    \item Three bands selection: three bands are chosen from the hyperspectral input data by the user based on experience, emphasizing spatial features that effectively represent the objects of interest.
    \item SAM segmentation: The selected three-band representation is fed to SAM, generating initial segmentation masks for all objects within the dataset.
    \item Grounding dino integration: grounding dino is applied to the three-band representation, using the user-defined language prompt that specifies either the desired objects to retain SAM's segmentation masks or the unwanted objects to exclude from SAM's segmentation masks.
    \item Intersection and exclusion filtering: based on the language prompt provided to grounding dino, an intersecting filtering or exclusion filtering step is selected on the segmentation masks, ensuring the preservation or elimination of specific objects, respectively.
    
    \item Final mask generation: from the resulting masks after the filtering steps, the objects of interest are retained in the original hyperspectral cube, while the remaining pixels are masked out.
\end{enumerate}
\begin{figure*}[t]
  \centering
  \includegraphics[width=\textwidth]{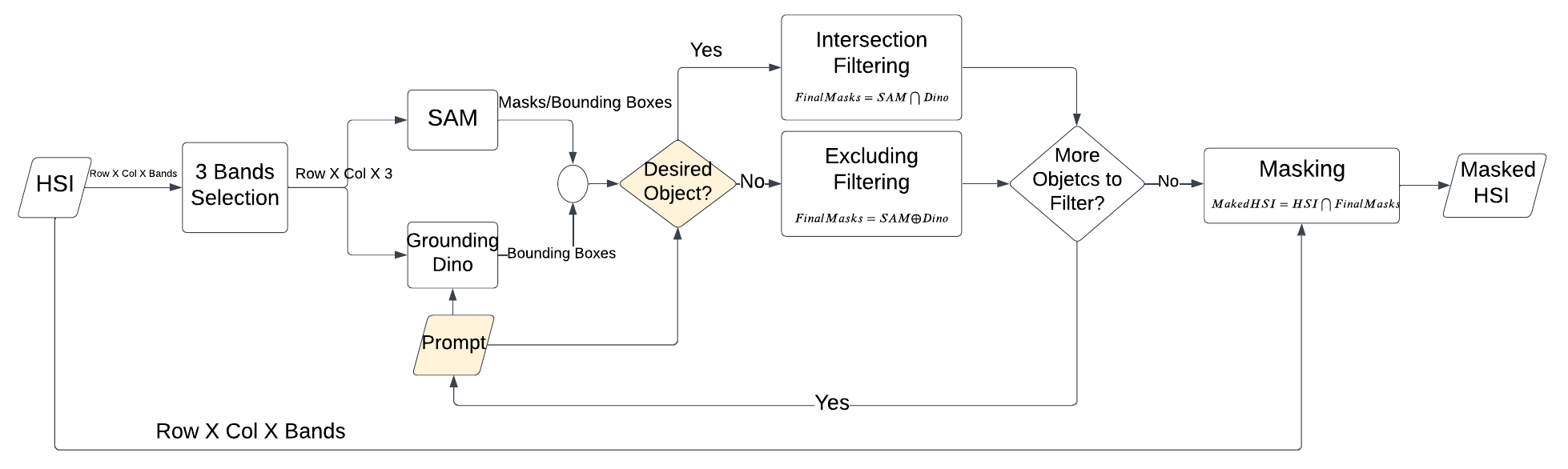} % Replace with your figure file name and extension
  \caption{Methodology Workflow.}
  \label{fig:flowchart}
\end{figure*}

Fig. \ref{fig:flowchart} presents the suggested method pipeline and flow of data. The spatial features quality of the three selected bands from the hyperspectral or multispectral data directly impacts the performance of SAM and grounding dino. To preserve SAM's desired masks, the intersection filtering step is selected, where bounding boxes from grounding dino and SAM are intersected based on the object of interest described in dino's language prompt. Alternatively, for removing undesired SAM's masks, the excluding filtering step is employed:  performing an XOR operation between SAM's predictions and dino's. This process may be repeated for multiple types of irrelevant objects. Finally, the masked hyperspectral image is obtained by projecting the final mask on the original hyperspectral cube.

\section{Experiments \& Results}
\label{sec:results}

This section presents three applications in which the proposed method was deployed, with the hyperparameters used to generate the results and a numerical evaluation relative to expert manually annotated ground truth. These results serve as compelling evidence for the effectiveness of the proposed method.

\subsection{Plastics Identification}

The identification of plastic materials is a crucial step for recycling and waste management, and it is the primary focus of RAMSES-4-CE, a project undertaken by our team at the Helios Lab, Helmholtz Institute Freiberg for Resource Technology. Hyperspectral imaging scans of the shredded plastics are conducted using SPECIM Aisa Fenix and RGB JAI. The suggested method is applied to the hyperspectral data, resulting in the generation of a final mask that is projected on the original hyperspectral cube to obtain the masked set. Once the masked hyperspectral data is obtained, a minimum wavelength mapping (MWM) \cite{hylite} technique and a decision tree are employed for hyperspectral data processing to identify the plastic material for each shredded plastic piece. The segmentation performance, relative to manually annotated ground truth, is summarized in Table \ref{table:eval}. The corresponding hyperparameters used in the process are presented in Table \ref{table:Hyperparameters}.
\begin{figure*}[t]
  \centering
  \includegraphics[width=\textwidth, scale=0.5]{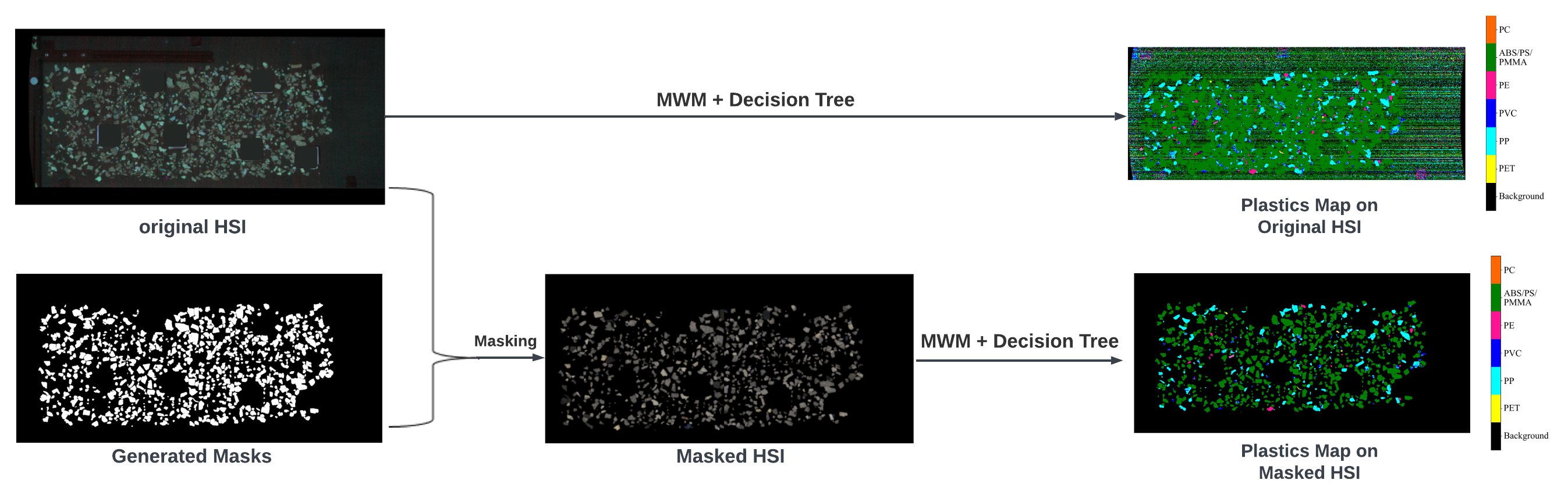} % Replace with your figure file name and extension
  \caption{Extracting Plastics Type in Material Streams.}
  \label{fig:plastic}
\end{figure*}

Fig. \ref{fig:plastic} presents the results flow at each step. The final mask (bottom-right) is acquired by deploying our method on the original hyperspectral data (false color representation, top-left) and using the masked hyperspectral data (middle) as input for the MWM and decision tree, which yields the final plastics-type map (top right). The efficiency  of the proposed method is demonstrated through improved denoising results achieved by utilizing a masked Hyperspectral Image (HSI) containing 145,258 vectors instead of the non-masked HSI (bottom right) with 863,360 vectors.

\subsection{Drill core scanning}
\begin{figure}
    \centering
    \includegraphics[width=1\linewidth]{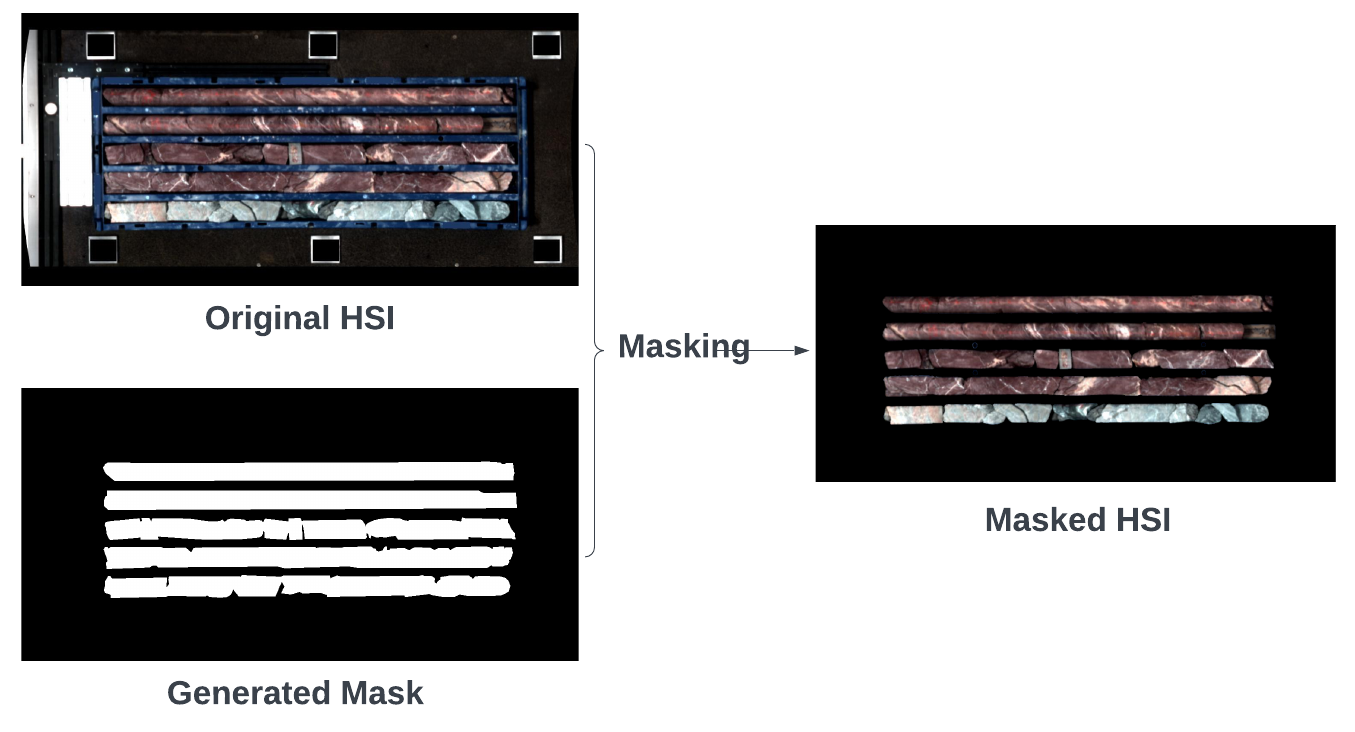}
    \caption{Extracting the Masked HSI in the Drill Core Scanning Application.}
    \label{fig:drillcore}
\end{figure}

Hyperspectral scanners are being increasingly used for mineral exploration, as they provide a rapid, non-destructive tool for gleaning otherwise cryptic mineralogical information from drillcores \cite{drillcore}. Drill cores are generally stored in open boxes, containing around 1-5 m of core, and scanned in succession to build up a database of many hundreds or thousands of individual scans. Each of these needs to be accurately masked to remove background and core tray pixels before mineralogical analyses can begin, a tedious and currently largely manual process. 
Fig. \ref{fig:drillcore} showcases the generation of the masked hyperspectral data (right) by projecting the method's generated mask (bottom-left) on the original hyperspectral image (top-left), effectively eliminating extraneous noise originating from the background. The utilization of a masked hyperspectral data significantly enhances the performance of subsequent processing techniques by processing 280,584 vectors in the masked hyperspectral cube instead of 727,000 vectors in the original one, ensuring optimal output quality. Table \ref{table:eval} demonstrates the highly accurate segmentation performance relative to the manually annotated ground truth. Table \ref{table:Hyperparameters} provides the hyperparameters used to generate the results. 

 \subsection{Litter monitoring}
\begin{figure}
    \centering
    \includegraphics[width=0.8\linewidth]{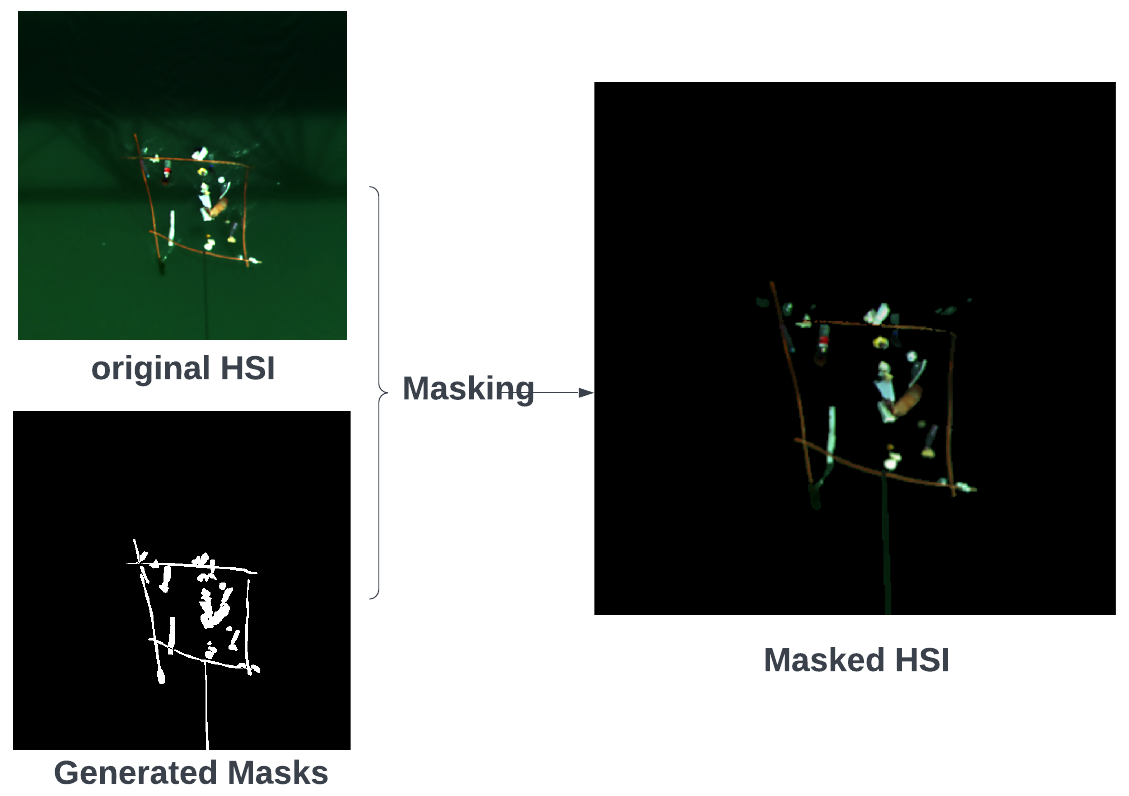}
    \caption{Extracting the Masked HSI in the Litter Monitoring Application.}
    \label{fig:3rd}
\end{figure}

The utilization of hyperspectral imaging, with its wide range of descriptive wavelengths in the visible and near-infrared region, is expected to enhance macro litter material identification in complex outdoor riverine scenarios \cite{litter}. 
Fig. \ref{fig:3rd} depicts one exemplary calibrated hypercube represented as the RGB composite (top left), and the generated mask (bottom left), which plays a crucial role in the processing workflow by reducing the 168,000 total vectors to 5,653 relevant ones (right). The application of masking techniques to identify relevant objects in the scene facilitates and accelerates subsequent analysis of the hyperspectral imagery. This analysis can involve techniques such as unmixing, feature extraction, and classification. Consequently, this approach effectively mitigates the influence of background noise (shadows and water), while reducing the computational requirements. The numerical segmentation performance relative to the manually annotated ground truth is demonstrated in table \ref{table:eval}, the high precision and recall scores are proofs of the proposed method's accuracy in masking out undesired backgrounds while preserving the desired regions, respectively. The used hyperparameters can  be seen in Table \ref{table:Hyperparameters}. The hyperparameter '\textit{C}' in the filtering steps is used as a margin for comparing coordination values between bounding boxes in pixels.'\textit{C}' should be chosen with trial and error based on the application's data spatial resolution.

\afterpage{
\begin{table*}[t]
    \centering
    \begin{tabular}{|c|c|c|c|}
        \hline
        \textbf{Application \textbackslash Metric} & \textbf{Precision} & \textbf{Recall} & \textbf{F1-Score} \\
        \hline
        \textbf{Plastics Identification} & 0.80 & 0.92 & 0.86 \\
        \hline
        \textbf{DrillCores Scanning} & 0.97 & 0.98 & 0.97 \\
        \hline
        \textbf{Litter Monitoring} & 0.77 & 0.87 & 0.82 \\
        \hline
    \end{tabular}
    \caption{The Proposed Method Segmentation Evaluation Metrics Relative to Manual Annotations on the Three Applications.}
    \label{table:eval}
\end{table*}

}
% Hyperparameters Tables
\afterpage{
\begin{table*}[t]
    \centering
    \begin{tabular}{|c|c|c|c|}
      \hline
\textbf{Hyperparameter} & \textbf{Shredded Plastics} & \textbf{Drill Core Scan} & \textbf{Litter Monitoring} \\
\hline
% SAM
\cellcolor{cyan!25} Points per Side & \cellcolor{cyan!25}256 & \cellcolor{cyan!25}128 & \cellcolor{cyan!25}128\\
\cellcolor{cyan!25} Points per Batch & \cellcolor{cyan!25}128 & \cellcolor{cyan!25}128 & \cellcolor{cyan!25}128\\
\cellcolor{cyan!25} Pred IOU Thresh & \cellcolor{cyan!25}0.7 & \cellcolor{cyan!25}0.7 & \cellcolor{cyan!25}0.8\\
\cellcolor{cyan!25} Crop \textit{n} Points Downscale Factor & \cellcolor{cyan!25}2 & \cellcolor{cyan!25}1 & \cellcolor{cyan!25}1\\
\hline

% GD
\cellcolor{green!25}Text Prompt & \cellcolor{green!25}"shredded piles of plastics" & \cellcolor{green!25}"cores" & \cellcolor{green!25}"object"\\
\cellcolor{green!25}Box Threshold & \cellcolor{green!25}0.4 & \cellcolor{green!25}0.5 & \cellcolor{green!25}0.1\\
\cellcolor{green!25}Text Threshold & \cellcolor{green!25}0.4 & \cellcolor{green!25}0.4 & \cellcolor{green!25}0.1\\
\hline
% Method
\cellcolor{orange!25}\textit{C} & \cellcolor{orange!25}15 & \cellcolor{orange!25}5 & \cellcolor{orange!25}5\\
      \hline
    \end{tabular}
  \centering
  \caption{Hyperparameters Used in the Proposed Method to Generate the Final Masks: SAM's Hyperparameters (Cyan), Grounding Dino (Green), the Filtering Steps (Orange).}
  
  \label{table:Hyperparameters}
\end{table*}
}
\section{Discussion}
\label{sec:Discussion}
Several alternative masking methods have been considered for comparison. Traditional computer vision techniques like alpha channel masking and thresholding necessitate manual parameter tuning for each image, especially when handling brightness variations. This presents a considerable challenge to the algorithm's reliability. Deep learning techniques have successfully addressed this issue by employing segmentation models such as DeepLab, Unet, and others. However, it is known that the performance of these models heavily depends on the quantity and quality of available training data. In contrast, our proposed method achieves comparable performance to manual expert masking, and it eliminates the need for training data and individual image parameter tuning.

\section{Conclusion}
\label{sec:Conclusion}
This work introduces a masking method that leverages various computer vision techniques to enhance the effectiveness of hyperspectral data processing pipelines. The method serves as a filtering approach that effectively masks out undesired backgrounds and unwanted objects in the hyperspectral cube, allowing the retainment of objects of interest only. By eliminating spectral vectors that introduce further noise, this approach enhances hyperspectral pre-processing tasks such as normalization and dimensionality reduction, as well as subsequent processing techniques such as classification.%, leading to enhanced performance.
The proposed method utilizes the SAM and the grounding dino zero-shot object detector, followed by intersection and exclusion filtering processes, resulting in a masked hyperspectral cube as the output. The masking generalization of the method is demonstrated in three applications: plastics identification, drill core scanning, and litter monitoring. 
The numerical evaluation demonstrates the proposed method's ability to mask regions of interest with high precision and recall proving its effectiveness as a promising alternative to laborious and time-intensive manual masking routines. Moreover, addressing the need for reliable solutions in scenarios where annotated training datasets for masking are scarce or non-existent.
\section{Acknowledgment}
The authors would like to thank EIT RawMaterials for funding the project ‘RAMSES-4-CE’ (KIC RM 19262). We acknowldge the European Regional Development Fund (EFRE) and the Land of Saxony for their funding of the computational equipment through the project ‘CirculAIre’.

\bibliographystyle{IEEEbib}
%\bibliography{paper}

\end{document}